\documentclass{article}
\usepackage{ijcai26}

\usepackage{times}
\usepackage{soul}
\usepackage{url}
\usepackage[hidelinks]{hyperref}
\usepackage[utf8]{inputenc}
\usepackage[small]{caption}
\usepackage{graphicx}
\usepackage{amsmath}
\usepackage{amsthm}
\usepackage{amssymb}
\usepackage{booktabs}
\usepackage{algorithm}
\usepackage{algorithmic}
\usepackage[switch]{lineno}
\usepackage{color}
\usepackage{multirow}

\usepackage[inline]{enumitem}
\usepackage{makecell}

\usepackage{amsthm}

\title{Uncertainty-Aware Motion Planning for Autonomous Driving 

in Mixed Traffic Environment}

\author{
    \affiliations
    Submission ID \#5638
}
\author{
Ming Cheng$^1$
\and
Hao Chen$^2$
\and
Ziyi Yang$^1$
\and
Ziluowen Luo$^1$
\and
Senzhang Wang$^1$\\
\affiliations
$^1$Central South University\\
$^2$City University of Macau\\
\emails
chengming@csu.edu.cn,
sundaychenhao@gmail.com,
8208240708@csu.edu.cn,
lzlwddl@csu.edu.cn,
szwang@csu.edu.cn
}

\begin{document}

\maketitle

\begin{abstract}
In mixed-traffic environments where autonomous and human-driven vehicles may co-exist, motion planning for autonomous vehicles requires anticipating the future behaviors of surrounding human drivers. Existing reinforcement learning-based methods generally directly incorporate the predicted human intents into the observation to enable a proactive planning. However, human intent is inherently uncertain due to the behavioral diversity, perception noise, and partial observability. Treating predicted intends as deterministic states can result in unsafe decisions for autonomous vehicles.
To address this problem, we propose \textbf{Uncertainty-Aware Motion Planning (UAMP)}, which incorporates uncertainty in human intent prediction for AV decision-making. Specifically, UAMP first introduces a proximity-aware uncertainty estimator to quantify the interaction-conditioned intent uncertainty and constructs an uncertainty-guided joint intent distribution over surrounding human-driven vehicles. Within this uncertainty set, UAMP further introduces Uncertainty-Calibrated Value Learning (UCVL) to correct value function learning biases arising from directly incorporating uncertain human intent predictions into the observation. Extensive experiments in various mixed-traffic scenarios show that UAMP significantly improves safety and driving comfort, while maintaining traffic efficiency compared with existing approaches. The code is released at~\url{https://anonymous.4open.science/r/UAMP-5638}. 
\end{abstract}

\section{Introduction}

\begin{figure}
  \centering
\includegraphics[width=0.5\textwidth]{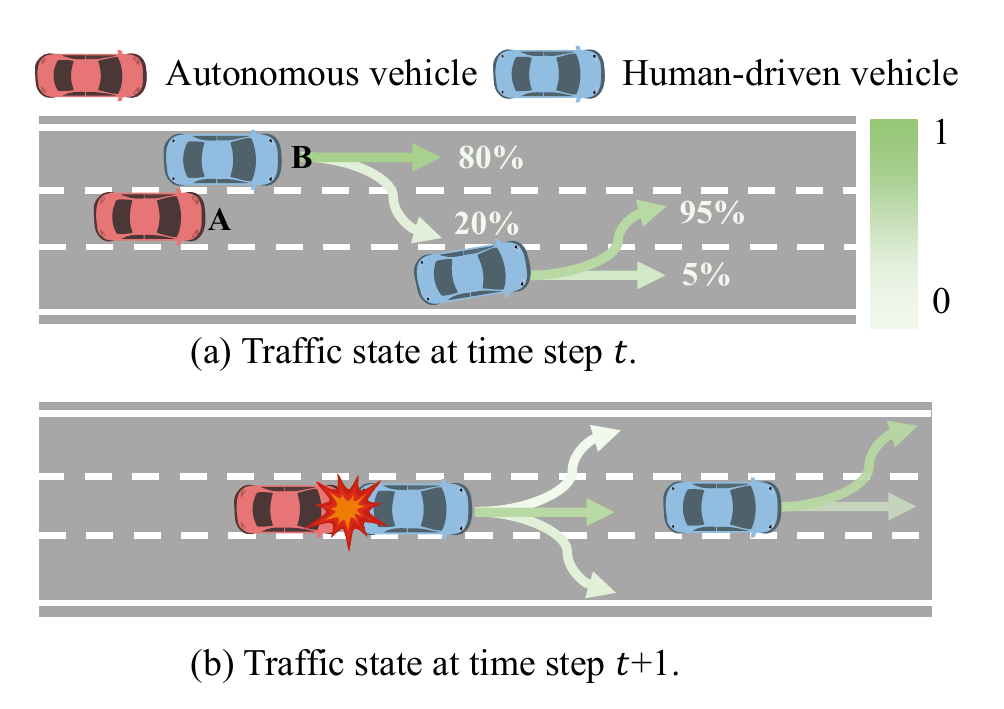} 
\caption{Illustration of the impact of intent prediction uncertainty on autonomous vehicles decision making in complex mixed-traffic environments.}
\label{intro}  
\end{figure}
Autonomous driving is being integrated into real-world transportation systems~\cite{papadoulis2019evaluating,kavas2021connected,dinneweth2022multi}. This leads to mixed-traffic environments where autonomous vehicles coexist with human-driven vehicles~\cite{hegyi2005model,yang2014control}. Since autonomous systems are still being gradually deployed and current rules and technology are not fully ready, mixed-traffic scenarios are expected to remain the dominant operating regime for autonomous driving systems. In such environments, autonomous vehicles must interact with surrounding human-driven vehicles to navigate safely and efficiently.
Therefore, decision making in autonomous driving systems must explicitly account for the behaviors of nearby human drivers~\cite{chandra2020forecasting}.

Existing methods for mixed-traffic decision-making can be broadly divided into rule-based approaches and learning-based approaches. Rule-based methods rely on predefined rules or simplified behavior models, such as rule engines and classical lane-changing models~\cite{rios2016survey}. However, these methods often require extensive manual design and tuning, and they may generalize poorly to diverse interaction patterns in mixed traffic scenarios~\cite{di2021survey}. Learning-based methods, especially reinforcement learning (RL), have attracted increasing research attention because they can learn decision policies from data and interaction experience~\cite{vinitsky2018benchmarks}. A common paradigm in RL-based mixed-traffic decision making is to explicitly predict the future behaviors or driving intents of surrounding human-driven vehicles and incorporate these predictions into the agent’s observation representation to inform decision making~\cite{shi2025predictive,guo2024mappo}. This paradigm allows the agent to reason about future behaviors beyond instantaneous traffic states, while treating predicted human intents as part of the state.

However, human intent prediction is inherently uncertain due to diverse driving behaviors, sensor noise, and perception errors~\cite{huang2022survey}.
When intent predictions are treated as deterministic state information and embedded into value learning, the value function bootstraps on potentially incorrect future assumptions.
As illustrated in~\autoref{intro}, at time step \(t\), autonomous vehicle A plans its maneuver based on a prediction that human-driven vehicle B will maintain its current lane and incorporates this prediction into state representation as well as value estimation.  
However, at time step \(t+1\), human-driven vehicle B unexpectedly changes lanes, deviating from the earlier prediction of autonomous vehicles A. 
Because autonomous vehicles A makes decisions based on the predicted intent of the human-driven vehicle B, it fails to respond instantly to the unexpected behavior of human-driven vehicle B, leading to a collision as shown in~\autoref{intro}.

Therefore, it is critically important to incorporate the uncertainty in human intent prediction into RL–based decision making for safer autonomous driving, but it faces the following two key challenges.
\textbf{ 1. Interaction-Conditioned Intent Uncertainty.}
In mixed-traffic environments, uncertainty arises from the behaviors of individual human-driven vehicles. However, autonomous vehicles decision-making requires modeling the uncertainty at the environment level based on interactions. Existing approaches lack an effective way to quantify such interaction-based uncertainty. \textbf{2. Uncertainty-Induced Value Bias.} Prediction errors in human intent can accumulate in value learning, leading to biased long-horizon estimations and overly aggressive policies. How to correct these value biases during reinforcement learning under intent uncertainty remains a challenging and underexplored research problem.

To address these challenges, we propose Uncertainty-Aware Motion Planning (UAMP), a framework for autonomous driving in mixed-traffic environments that incorporates predicted uncertainty in human driving intent into decision making. Specifically, UAMP first constructs an ego-centric interaction graph and predicts the intents of surrounding human-driven vehicles. Based on the predicted intents, we develop proximity-aware uncertainty estimator to a quantifies the uncertainty of each surrounding vehicle and aggregate it through interactions into an environment-level uncertainty radius. Furthermore, we constructs an uncertainty-guided joint intent distribution to capture the collective behavioral ambiguity of multiple interacting vehicles. Using this joint intent distribution, UAMP applies Uncertainty-Calibrated Value Learning (UCVL) to adjust value estimates within the uncertainty radius, enabling more conservative decision-making under high uncertainty.

In summary, our contributions are as follows:
\begin{itemize}
    \item We propose Uncertainty-Aware Motion Planning (UAMP), the first framework to explicitly incorporate predicted human intent uncertainty into RL–based motion planning for autonomous driving.

    \item We define proximity-aware intent uncertainty in traffic scenarios and introduce Uncertainty-Calibrated Value Learning (UCVL), which uses this uncertainty to correct value estimates and enable a safer decision making.

    \item We validate the effectiveness of our approach through extensive multi-lane mixed-traffic experiments, showing that UAMP significantly improves driving safety and comfort while maintaining traffic efficiency by comparison with baseline methods .
\end{itemize}

\section{Preliminaries}

We formulate the decision-making process of autonomous vehicles in mixed-traffic environments as a decentralized partially observable Markov decision process (Dec-POMDP),
denoted by
$\mathcal{M} = \langle \mathcal{I}, \mathcal{S}, \mathcal{A}, \mathcal{P}, \mathcal{O}, \mathcal{R}, \gamma \rangle$,
where $\mathcal{I} = \{1, \dots, i\}$ denotes the set of autonomous vehicle agents, $\mathcal{S}$ represents the set of all possible environment states, $\mathcal{A}$ is the set of joint actions, $\mathcal{P}$ is the transition probability function, $\mathcal{O}$ denotes the set of possible observations for each agent, $\mathcal{R}$ is the reward function assigning feedback to states and actions, and $\gamma \in [0,1)$ is the discount factor. At each time step, autonomous vehicle agents select actions based on their own local observations.

\paragraph{Observation.}
At each time step $t$, each autonomous vehicle agent $i$ receives a local observation
$o_t^i \in \mathcal{O}$ that provides an ego-centric representation of the surrounding traffic environment.
The observation comprises the state of the ego vehicle and the states of vehicles within a fixed perception radius.

The ego vehicle state includes its current speed, longitudinal acceleration, lane index, and local traffic density.
For surrounding vehicles, the autonomous vehicle observes vehicles located within a fixed perception range.
For each neighboring vehicle, the observation contains relative position, relative speed, relative acceleration, and lane index.

\paragraph{Action Space.}
Based on its observation, each autonomous vehicle selects an action $a_t^i \in \mathcal{A}$.
 from a discrete action space. The action space consists of five motion primitives: \textit{maintaining the current speed, accelerating, decelerating, changing lane to the left, and changing lane to the right}.

\paragraph{Reward Function.}
At each time step $t$, each autonomous vehicle agent receives an environment reward $r_t \in \mathcal{R}$ that encodes multiple driving objectives.
The reward is formulated as a weighted combination of terms reflecting efficiency, safety, and ride comfort.
Specifically, the reward encourages efficient travel, penalizes collisions with large negative values,
and discourages uncomfortable maneuvers through penalties on excessive acceleration and jerk. The overall goal of all agents is to maximize the expected cumulative discounted reward, defined as follows:
\begin{equation}
J = \mathbb{E}\Bigg[\sum_{t=0}^{T} \gamma^t \sum_{i \in \mathcal{I}} r_t^i \Bigg].
\end{equation}

\section{Methodology}
\begin{figure*}
  \centering
\includegraphics[width=1\textwidth]{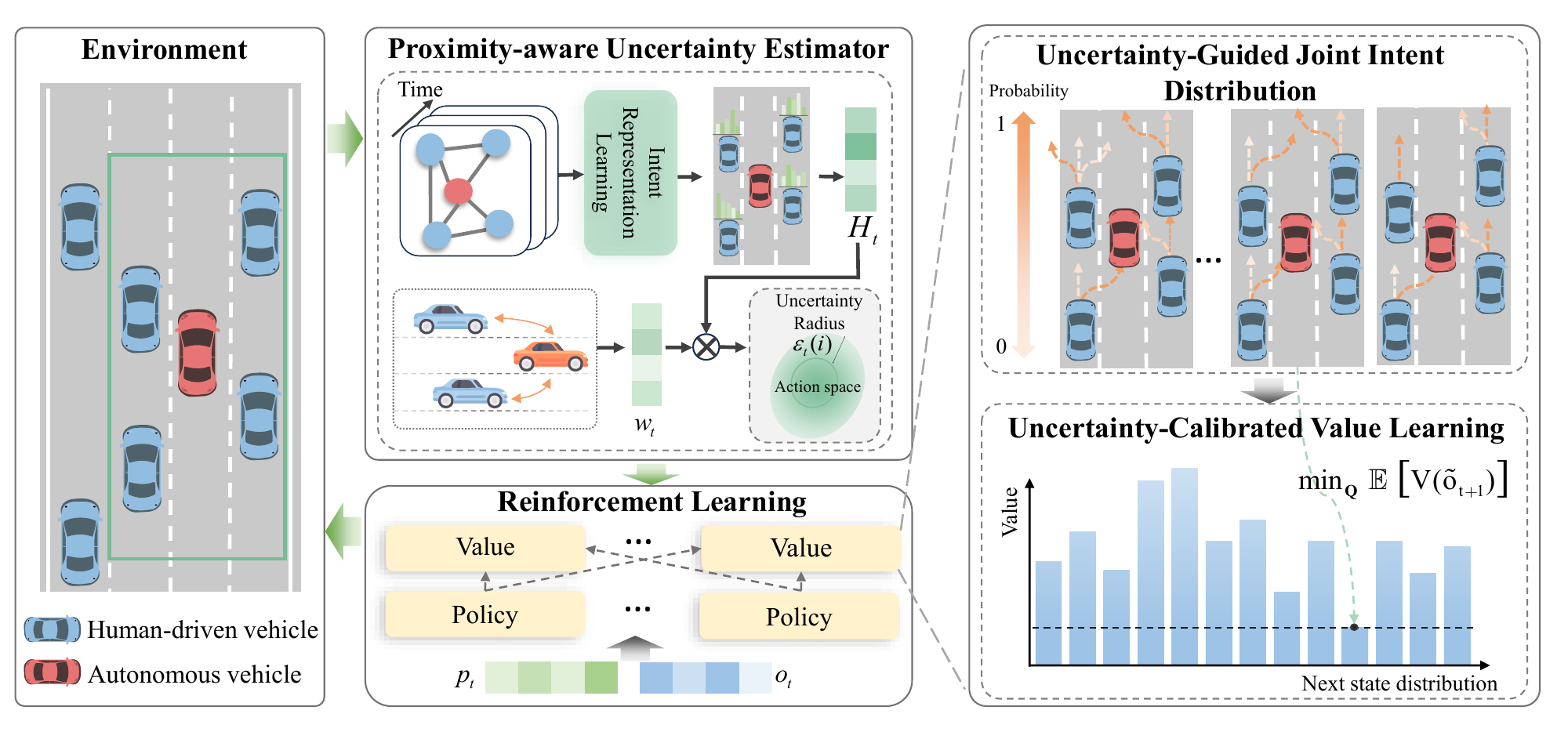} 
\caption{Overview of the UAMP framework. UAMP estimates proximity-aware uncertainty in human intent prediction from an ego-centric perspective and incorporates this uncertainty into long-horizon decision making through uncertainty-calibrated value learning.}
\label{framework}  
\end{figure*}
\autoref{framework} illustrates the proposed UAMP framework.
First, the proximity-aware uncertainty estimation module takes observations from the traffic environment to predict surrounding human driving intents and quantify interaction-conditioned uncertainty from an ego-centric perspective.
Based on the estimated uncertainty, the uncertainty-guided joint intent distribution module samples plausible joint intents within the corresponding uncertainty radius.
These sampled joint intent distributions are used to calibrate the value function in reinforcement learning, enabling value estimation that accounts for intent uncertainty.
Next, we introduce the proposed UAMP framework in detail.

\subsection{Proximity-aware Uncertainty Estimator}

Proximity-aware uncertainty modeling aims to capture the uncertainty associated with inferred future intent representations of surrounding human-driven vehicles. This uncertainty is influenced by evolving interaction patterns and partial observability in mixed-traffic scenarios, leading to variations in the confidence of intent prediction over time. Based on these considerations, we design an uncertainty estimator to model intent prediction uncertainty for value learning.

\paragraph{Intent Representation Learning.}
At each time step, the local traffic environment faced by the ego agent evolves as surrounding vehicles enter  or leave.
As shown in~\autoref{framework}, we construct an ego-centric interaction graph \( G_t = (V_t, E_t) \) over vehicles within the perception range of the ego agent.
Each node $v \in V_t$ corresponds to a human-driven vehicle. We focus on human-driven vehicles because their behaviors are unknown and need to be explicitly inferred, whereas other autonomous vehicles follow the same policy as the ego agent and do not require separate intent modeling.
Edges in \( E_t \) encode local interaction relationships, such as spatial proximity or lane-level relevance.

To capture the temporal behavior patterns, we summarize the observation history of each surrounding vehicle into a latent state using a recurrent encoder updated online at each time step. For each vehicle $v$, the latent state is updated as,
\begin{equation}
\mathbf{h}_t^v, \mathbf{c}_t^v = \mathcal{F}_\psi\left( \mathbf{u}_t^v, \mathbf{h}_{t-1}^v, \mathbf{c}_{t-1}^v \right),
\end{equation}
where $\mathbf{u}_t^v$ denotes the local observation features at time $t$ and $\mathbf{h}_t^v$ aggregates past observations.


To incorporate interaction effects, the latent states are further refined using a graph-based interaction encoder, and intent representations are
inferred as:

\begin{equation}
\bar{\mathbf{h}}_t^v
=
\sigma\!\left(
\mathbf{W}_1 \mathbf{h}_t^v
+
\sum_{u \in \mathcal{N}_v}
\alpha_{vu}
\mathbf{W}_2 \mathbf{h}_t^u
\right),
\end{equation}
\begin{equation}
p_t^v(\mathcal{A}) = \pi_\eta\left(\bar{\mathbf{h}}_t^v\right),
\end{equation}
where $\mathbf{W}_1$ and $\mathbf{W}_2$ are learnable weight matrices, $\sigma(\cdot)$ denotes a nonlinear activation function, and $\alpha_{vu}$ denotes normalized attention weights over neighbors $u \in \mathcal{N}_v$, $\mathcal{N}_v$ denotes the set of neighboring vehicles of $v$ in the interaction graph, $\pi_\eta(\cdot)$ denotes a parametric intent decoder, and $p_t^v(\mathcal{A})$ denotes the categorical probability distribution over the discrete action space $\mathcal{A}$.

\paragraph{Proximity-Aware Uncertainty.}

We model intent prediction uncertainty as an ego-centric and proximity-aware quantity that represents the ambiguity in the future intents of surrounding human-driven vehicles at each time step from the perspective of the autonomous agent.

For each surrounding human-driven vehicle \( v \), we measure the uncertainty of its inferred intent via entropy as follows:
\begin{equation}
    H_t^v = - \sum_{a \in A} p_t^v(a)\log p_t^v(a),
\end{equation}
where a higher entropy indicates a larger ambiguity in the predicted future behavior of vehicle \( v \).

To account for the varying influence of surrounding vehicles on the ego agent, we introduce an ego-centric proximity-based weighting mechanism.
Let \( d_t(i, v) \) denote the Euclidean distance between the ego agent \( i \) and vehicle \( v \) at time \( t \).
The proximity weight is defined as:
\begin{equation}
    w_t(i, v) = \frac{\exp\left(-\alpha d_t(i, v)\right)}{\sum_{u \in \mathcal{N}_i} \exp\left(-\alpha d_t(i, u)\right)},
\end{equation}
where \( \alpha > 0 \) controls the spatial decay rate and \( \mathcal{N}_i \) denotes the set of vehicles within the ego agent’s vicinity.

The overall proximity-aware intent prediction uncertainty for agent \( i \) is obtained by aggregating vehicle-level uncertainties using the ego-centric proximity weights as follows:
\begin{equation}
    \varepsilon_t^{i} = \sum_{v \in \mathcal{N}_i} w_t(i, v) H_t^v.
\end{equation}
The resulting uncertainty measure \( \varepsilon_t^{i} \) characterizes the uncertainty in the predicted future intents of surrounding human-driven vehicles that is most relevant to the ego agent, and is subsequently employed as a sampling radius to bound and guide the generation of joint intent distributions for downstream decision making.

\subsection{Uncertainty-Guided Joint Intent Distribution}

To account for the uncertainty in the behaviors of surrounding human-driven vehicles, this subsection constructs an uncertainty-guided joint intent distribution.

At each time step $t$, the intent prediction module outputs a probabilistic
intent distribution $p_t^v$ for each human-driven vehicle $v$ within the ego
agent perceivable neighborhood $\mathcal{N}_i$.
Each distribution captures the predicted driving intent of the corresponding individual vehicle.

To reason about the joint influence of surrounding vehicles on the ego agent’s
decision making, we aggregate the individual intent predictions \( \{ p_t^v \}_{v \in \mathcal{N}_i} \) and construct
an ego-centric joint intent distribution.
Specifically, assuming the conditional independence given the current observation,
the nominal joint distribution is defined as:
\begin{equation}
\mathbf{P}_t^{i}(\mathbf{a}_t^{hdv}
)
=
\prod_{v \in \mathcal{N}_i} p_t^v(a^v),
\end{equation}
where $\mathbf{a}_t^{hdv}=(a_t^v)_{v\in \mathcal{N}_i}$ denotes the joint intent configuration of surrounding human-driven vehicles in $\mathcal{N}_i$.

However, due to partial observability, limited sensing range, and complex
multi-agent interactions, the true joint behaviors of surrounding vehicles
cannot be perfectly inferred.
As a result, the nominal joint intent distribution
$\mathbf{P}_t^{(i)}$ is inherently uncertain and should not be treated as fully
reliable.

To explicitly account for this uncertainty, we leverage the proximity-aware
intent uncertainty measure $\varepsilon_t^{i}$ and define a set of uncertainty-guided joint intent distribution around the prediction $\mathbf{P}_t^{i}$.
We construct the uncertainty set as follows:
\begin{equation}
\mathcal{U}_t^{i} =
\left\{
\mathbf{Q}
\;\middle|\;
\frac{1}{2}
\left\|
\mathbf{Q} - \mathbf{P}_t^{i}
\right\|_1
\le
\varepsilon_t^{i}
\right\},
\end{equation}
where $\mathbf{Q}$ denotes an alternative joint intent distribution. When intent predictions are confident, the uncertainty set concentrates around the nominal joint distribution; under high uncertainty, a broader set of joint intents must be considered.

\subsection{Uncertainty-Calibrated Reinforcement Learning}

Incorporating predictive intent information into an agent’s observations may bias value estimation under uncertain intent predictions. We therefore propose an uncertainty-calibrated value learning method that adjusts value propagation based on proximity-aware intent uncertainty.

\paragraph{Uncertainty-Calibrated Value Learning}

In order to reason about the future behaviors of surrounding vehicles, the ego agent augments its physical observation with predicted intent information. Specifically, at time step $t$, the ego agent receives an augmented state as follows:

\begin{equation}
\hat{o}_t^{i} = (o_t^{i}, \{ p_t^v \}_{v \in \mathcal{N}_i}),
\label{eq:9}
\end{equation}
where \( o_t^i \) denotes the state returned by the environment, and \( p_t^v \) represents the predicted intent distribution of each surrounding vehicle \( v \). 
While augmenting observations with intent information enables anticipation of future interactions, directly incorporating the predicted intent distribution \( p_t^v \) as part of the state may introduce bias in the value estimation. This is because the predicted intent distribution \( p_t^v \) is uncertain, and the true underlying distribution \( q_t^v \) may differ significantly from the prediction. The bias in the value estimation can be expressed as:

\begin{equation}
\Delta V = \left| \mathbb{E}_{p_t^v} \left[ V(\hat{o}_t^i) \right] - \mathbb{E}_{q_t^v} \left[ V(\hat{o}_t^i) \right] \right|.
\label{eq:10}
\end{equation} 
where \( p_t^v \) be the predicted intent distribution for vehicle \( v \), and \( {q}_t^v \) be the true intent distribution. By substituting the definition of the augmented observation from Eq.~\ref{eq:9} and applying the standard total-variation bound, we obtain (the proof is deferred to Appendix~\ref{app:proof}):

\begin{equation}
\Delta V \leq 2\, \varepsilon_t^i(V_{\max} - V_{\min}) ,
\end{equation}
where \(V_{\max}, V_{\min}\) are the bounds of the value function. 

This inequality shows that the magnitude of the bias is directly proportional to the uncertainty radius \(\varepsilon_t^i\). A larger uncertainty radius leading to a larger potential error in value estimation.
Based on this observation, we perform value backup over the uncertainty set of joint intent distributions. Accordingly, the uncertainty-calibrated Temporal Difference (TD) target is defined as follows:
\begin{equation}
y_t^i = r_t^i + \gamma \min_{\mathbf{Q} \in \mathcal{U}_t^{i}} \mathbb{E}_{\mathbf{Q}} \left[ V(\widetilde{o}_{t+1}^i) \right],
\end{equation}

\begin{equation}
\mathbb{E}_{\mathbf{Q}} \left[ V(\widetilde{o}_{t+1}^i) \right] = \sum_{\mathbf{a}^{\mathrm{human}} \in \mathcal{A}} \mathbf{Q}\left( \mathbf{a}^{\mathrm{hdv}} \right) V(\widetilde{o}_{t+1}^i),
\end{equation}
where the next augmented observation \( \widetilde{o}_{t+1} \) depends jointly on the ego agent’s action and the future behaviors of surrounding vehicles:
\begin{equation}
\widetilde{o}_{t+1}^i = f\left(\hat{o}_t^i, \mathbf{a}_t^{\mathrm{auto}}, \mathbf{a}_t^{\mathrm{hdv}}\right),
\end{equation}
where \( \mathbf{a}_t^{\mathrm{hdv}} \) denotes the joint future actions of nearby human-driven vehicles, and \( f \) represents the environment transition. 
In implementation, the minimum expected value over \( \mathcal{U}_t^i \) is approximated via sampling-based evaluation of joint actions.

This formulation yields a conservative value estimation when intent prediction uncertainty is high, while recovering the standard Bellman backup as \( \varepsilon_t^{i} \rightarrow 0 \).

\renewcommand{\arraystretch}{1.4}
\begin{table*}[htbp]
\centering
\caption{Performance comparison of different methods under varying traffic scenarios.}
\label{tab:driving_style_comparison}
\resizebox{\textwidth}{!}{
\begin{tabular}{lcccccccccccc}
\toprule
\multirow{2}{*}{Method} 
& \multicolumn{4}{c}{2-Lane Configuration} 
& \multicolumn{4}{c}{3-Lane Configuration} 
& \multicolumn{4}{c}{4-Lane Configuration} \\
\cmidrule(lr){2-5} \cmidrule(lr){6-9} \cmidrule(lr){10-13} 
& \makecell{ $V_{\text{avg}}$$\uparrow$}
& \makecell{$T_{\text{travel}}$$\downarrow$}
& \makecell{$J_{\text{acc}}$$\downarrow$}
& \makecell{$R_{\text{success}}$$\uparrow$}

& \makecell{$V_{\text{avg}}$$\uparrow$}
& \makecell{$T_{\text{travel}}$$\downarrow$}
& \makecell{$J_{\text{acc}}$ $\downarrow$}
& \makecell{$R_{\text{success}}$$\uparrow$}

& \makecell{ $V_{\text{avg}}$$\uparrow$}
& \makecell{ $T_{\text{travel}}$$\downarrow$}
& \makecell{$J_{\text{acc}}$$\downarrow$} 
& \makecell{$R_{\text{success}}$$\uparrow$}\\
\midrule
MVCM 
& 13.236 & 38.400 & 0.551 & 1.000
& 14.739 & 34.700 & 0.368 & 1.000
& 20.529 & 25.500 & 0.617 & 1.000 \\

rMAPPO 
& 7.993  & 58.900 & 0.299 & 0.850 
& 15.135 & 49.400 & 0.658 & 0.910
& 21.443 & 24.400 & 0.855 & 1.000 \\

IPPO 
& 14.514 & 41.200 & 0.243 & 1.000  
& 13.373 & 45.400 & 0.306 & 0.960
& \textbf{22.579} & 23.500 & 0.869 & 1.000 \\

P-DRL 
& 11.894 & 53.333 & 0.516 & 0.940  
& 14.505 & 38.200 & 0.273 & 0.980 
& 18.955 & 24.200 & 0.832 & 0.980 \\

MAPPO-PIS 
& 11.840 & 62.500 & 0.422 & 0.910  
& 12.301 & 55.200 & 0.209 & 0.960
& 20.623 & 27.800 & 0.632 & 1.000 \\
\midrule
\textbf{Ours} 
& \textbf{15.242} & \textbf{38.400} & \textbf{0.143} & \textbf{1.000}
& \textbf{17.472} & \textbf{31.600} & \textbf{0.203} & \textbf{1.000}
& 20.286 & \textbf{21.300} & \textbf{0.567}  & \textbf{1.000}\\
\bottomrule
\end{tabular}
}
\end{table*}

\paragraph{Policy Optimization with Calibrated Values}

We employ Proximal Policy Optimization (PPO) to train the model, with value updates guided by the uncertainty-calibrated TD targets described earlier. The loss function is defined as:

\begin{equation}
\begin{aligned}
\mathcal{L}(\theta, \phi)
=
&\mathbb{E}
\Big[
 \min \big(
r_{t}(\theta)\mathbf{A}_{t},
\text{clip}(r_{t}(\theta), \\
& 1-\epsilon, 1+\epsilon)\mathbf{A}_{t}
\big) + \left(
V_\phi(\hat{o}_t, i) - R_{t}
\right)^2
\Big],
\end{aligned}
\end{equation}

\begin{equation}
r_{t}(\theta) = \frac{\pi_\theta(a_{t} \mid \hat{o}_{t})}{\pi_{\theta_{\text{old}}}(a_{t} \mid \hat{o}_{t})},
\end{equation}
where $R_t$ is the cumulative return at time $t$, \( \mathbf{A}_{t} \) is the advantage function, defined as:


\begin{equation}
\mathbf{A}_{t} = \sum_{l=0}^{\infty} (\gamma \lambda)^l \delta_{t+l},
\quad
\delta_t = y_t^i - V_\phi(\hat{o}_t^i),
\end{equation}
where $\gamma \in (0,1)$ denotes the discount factor, $\lambda \in [0,1]$ controls the temporal smoothing of advantage estimation.

\section{Experiments}
\subsection{Experimental Setup}
\textbf{Simulation Environment.}
We conduct our experiments in the microscopic traffic simulator SUMO~\cite{lopez2018microscopic}, which provides high-fidelity modeling of vehicle behaviors and interactive traffic dynamics through the TraCI interface. 
As a widely adopted platform in autonomous driving and traffic research, SUMO enables realistic evaluation of motion planning and decision-making algorithms in complex traffic scenarios~\cite{9351818}.

\paragraph{Traffic Scenarios.}
We consider mixed-traffic environments consisting of both autonomous vehicles and human-driven vehicles. We consider three traffic scenarios based on road configurations:  \emph{2-Lane}, \emph{3-Lane}, and \emph{4-Lane} roads.  To simulate realistic behavioral uncertainty, human-driven vehicles follow heterogeneous driving styles, including \emph{aggressive}, \emph{normal}, and \emph{cautious}, with configurable proportions. 

Detailed configurations of the traffic scenarios setting are provided in Appendix~\ref{app:scenario_settings}.

\paragraph{Baselines.}We compare the proposed UAMP framework with a set of representative car-following and multi-agent reinforcement learning baselines, covering both ruled-based approaches and learning-based methods. 
The model-based baseline includes MVCM~\cite{wang2021mvcm}, while the learning-based baselines include IPPO~\cite{de2020independent}, rMAPPO~\cite{yu2022surprising}, P-DRL~\cite{shi2025predictive}, and MAPPO-PIS~\cite{guo2024mappo}. Further implementation details of the baseline methods are provided in Appendix~\ref{app:Baseline Details}.

\paragraph{Evaluation Metrics.}
Following prior work~\cite{guo2024mappo}, we evaluate performance using four metrics that capture efficiency, safety, and ride comfort.

\textit{Episodic Average Speed ($V_\text{avg}$).}
$V_\text{avg}$ measures the average driving speed of autonomous vehicles in an episode.

\textit{Average Travel Time ($T_\text{travel}$).}
$T_{\text{travel}}$ measures the average number of time steps required for the ego vehicle to complete the task and reach its goal.

\textit{Acceleration Cost ($J_{\text{acc}}$ ).}
\( J_{\text{acc}} \) quantifies ride comfort by measuring the magnitude of longitudinal acceleration fluctuations.
It is defined as:
\begin{equation}
J_{\text{acc}}
=
\frac{1}{|\mathcal{I}|}
\sum_{i \in \mathcal{I}}
\sqrt{
\frac{1}{T_i}
\sum_{t=1}^{T_i}
\left(\alpha_t^{i}\right)^2
},
\end{equation}
where $\mathcal{I}$ denotes the set of autonomous vehicles, $\alpha_t^{i}$ is the longitudinal acceleration of vehicle $i$ at time step $t$, and $T_i$ is its active duration.

\textit{Success Rate($R_{\text{success}}$).}
$R_{\text{success}}$ is the overall proportion of episodes in which the autonomous vehicle completes its task without collision.

\subsection{Main Result Analysis}

The quantitative comparison result under different network configurations are summarized in~\autoref{tab:driving_style_comparison}. From the table, one can derive the following observations.


\paragraph{Comparison with Rule-Based Method.}
MVCM serves as a representative rule-based car-following baseline that relies on deterministic control logic.
As shown in~\autoref{tab:driving_style_comparison}, MVCM achieves high success rates across all network settings, reflecting its robustness under simplified assumptions.
However, this robustness comes at the cost of limited adaptability.
MVCM exhibits lower average speed and noticeably higher acceleration fluctuations compared to UAMP, indicating less efficient and less smooth driving behavior.
In contrast, UAMP adapts its decisions based on uncertainty-aware value estimation, enabling smoother interactions and improved safety without sacrificing efficiency.


\paragraph{Comparison with RL-Based Methods.}
Compared with RL-based methods, UAMP achieves consistent and substantial performance improvements across multiple metrics by explicitly modeling interactions with human-driven vehicles and their future driving intents. In the 2-Lane scenario, where all methods achieve a success rate of $1.0$, UAMP reduces the average travel time by $34.8\%$ compared to rMAPPO. Similarly, in the 3-Lane configuration, UAMP shortens travel time by $36.0\%$ relative to rMAPPO while maintaining the same success rate, indicating that RL-based methods take substantially longer to reach the destination under comparable interaction scenarios. In terms of ride comfort, UAMP also demonstrates consistent improvements, reducing acceleration jerk by $41.2\%$ compared to IPPO in the 2-Lane scenario and by $34.7\%$ in the 4-Lane scenario, which indicates smoother driving behavior. These improvements do not come at the cost of efficiency, as UAMP maintains comparable or higher average speed across all configurations.

\paragraph{Comparison with Deterministic Prediction-Based RL Methods.}

Prediction-informed RL baselines such as P-DRL and MAPPO-PIS incorporate human driving predictions into decision making but treat these predictions as deterministic, ignoring their inherent uncertainty. Consequently, prediction errors may misguide the learned policy, leading to reduced success rates and degraded ride comfort. Across different traffic configurations, UAMP consistently achieves higher success rates than both P-DRL and MAPPO-PIS. In the 2-Lane scenario, UAMP improves the success rate by $6.4\%$ over P-DRL and by $9.9\%$ over MAPPO-PIS, while in more complex settings UAMP maintains a $2.0\%$ higher success rate than P-DRL in both the 3-Lane and 4-Lane configurations. In terms of ride comfort, UAMP yields substantial improvements, reducing acceleration jerk by $72.3\%$ in the 2-Lane scenario and by $31.9\%$ in the 4-lane scenario compared to P-DRL, and consistently achieving lower jerk than MAPPO-PIS across all configurations. Importantly, these improvements in success rate and ride comfort are obtained without sacrificing efficiency, as UAMP maintains comparable or higher average speed across all configurations.

Overall, the results demonstrate that UAMP effectively balances efficiency, safety, and comfort across varying levels of traffic complexity.

\begin{figure}[ht]
  \centering
\includegraphics[width=0.5\textwidth]{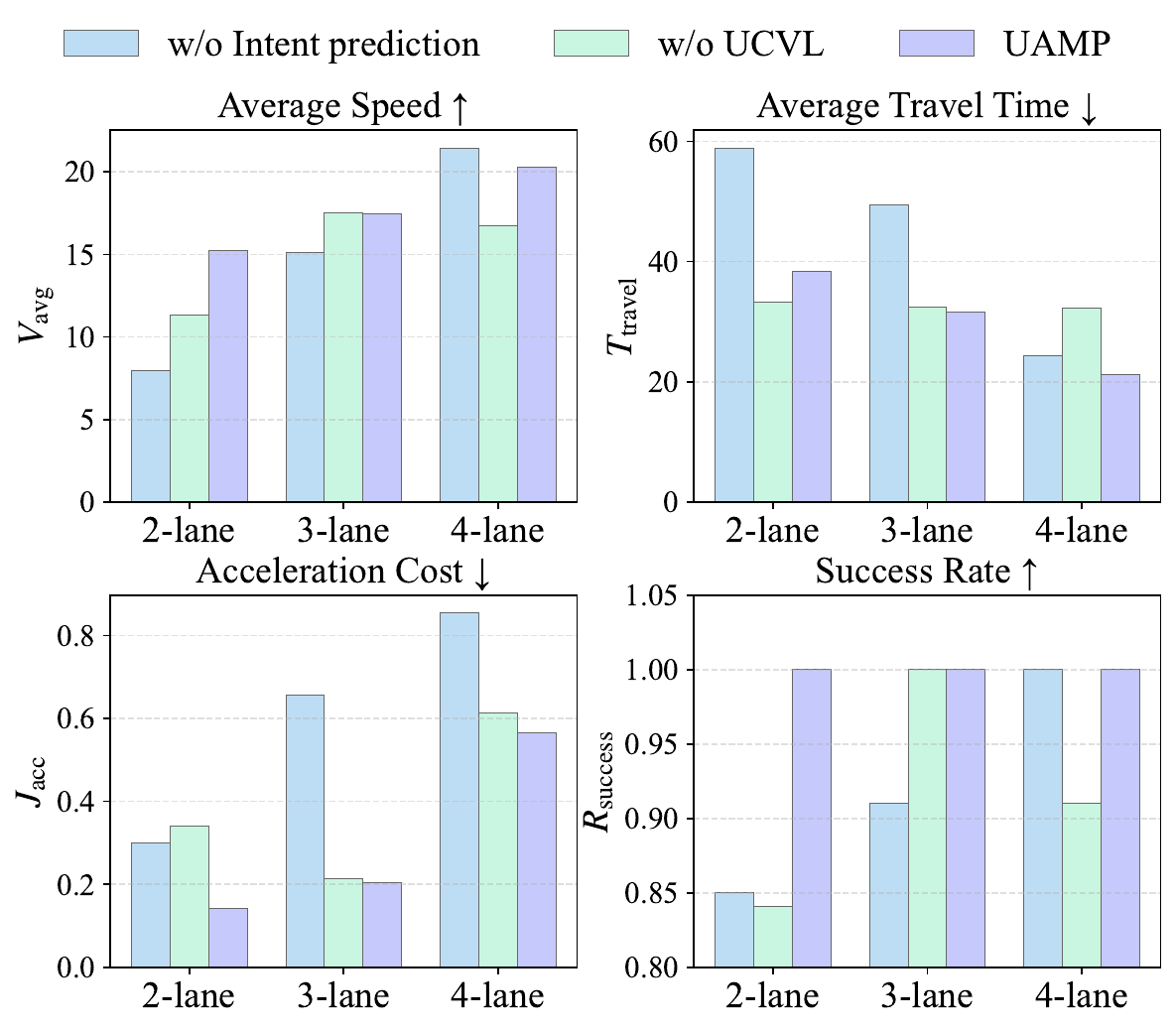} 
\caption{Ablation study results between UAMP with its two variants across three experimental scenarios..}
\label{ab_1}  
\end{figure}

\subsection{Ablation Study}

To investigate the contribution of each key component in UAMP, we conduct an ablation study with the following variants:
\begin{enumerate*}[label=(\roman*)]
    \item \textit{w/o Intent Prediction}: removes the intent prediction module and relies solely on instantaneous observations of surrounding vehicles.
    \item \textit{w/o UCVL}: removes UCVL and applies standard temporal-difference learning.
\end{enumerate*}

The results consistently show that introducing intent prediction leads to higher driving speeds and shorter arrival times across all scenarios, indicating that access to future intention information enables more anticipation and efficient decision making, as illustrated in~\autoref{ab_1}. However, when prediction uncertainty is ignored, this efficiency gain is accompanied by a decrease in the proportion of successfully arrived vehicles and a noticeable degradation in driving comfort, suggesting that agents rely on potentially inaccurate predictions and adopt overly aggressive behaviors. In contrast, when prediction uncertainty is explicitly considered, the model is able to largely preserve the efficiency benefits brought by intention prediction while significantly reducing collision rates and improving driving comfort. This demonstrates that uncertainty-aware decision making helps balance efficiency and safety.

\subsection{Performance under Different Scenario Settings}

To further evaluate the robustness of UAMP under uncertainty conditions, we compare UAMP with both rMAPPO and MVCM across multiple mixed-traffic scenarios.

\paragraph{Impact of Traffic Density.}
As traffic density increases, vehicle interactions become more frequent and human driving behaviors exhibit higher uncertainty.
As shown in~\autoref{tab:scene_density}, UAMP consistently outperforms rMAPPO and MVCM across low, medium, and high density scenarios.
In high-density traffic, UAMP improves efficiency and comfort relative to rMAPPO, reducing travel time by about $10.9\%$ and lowering acceleration cost by about $23.6\%$, indicating safer and smoother behavior under congestion.
These results suggest that explicitly modeling intent uncertainty enables UAMP to remain stable as interaction complexity increases.

\begin{figure}
  \centering
\includegraphics[width=0.5\textwidth]{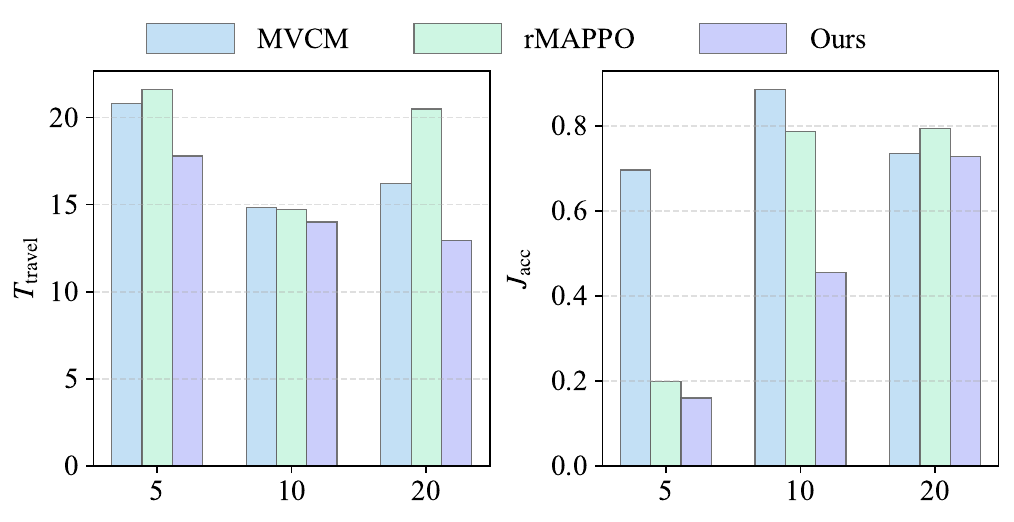} 
\caption{Comparison of performance metrics under varying autonomous vehicle penetration in the 3-Lane scenario.}

\label{ex_3}  
\end{figure}

\paragraph{Impact of Autonomous Vehicles Penetration.}
As shown in~\autoref{ex_3}, we further examine robustness under varying autonomous vehicles penetration levels.
When autonomous vehicles penetration is low, each autonomous vehicle interacts with more human-driven vehicles, resulting in higher behavioral uncertainty.
Under this setting, UAMP achieves shorter travel time and lower acceleration cost than rMAPPO, demonstrating stronger robustness.
As autonomous vehicles penetration increases, the gap in travel time becomes smaller; however, UAMP consistently maintains lower acceleration variance, indicating more stable control behavior.

\paragraph{Impact of Human Driving Style Distribution.}
Finally, we evaluate performance under different human driving style distributions. As shown in~\autoref{tab:hv_style} As human driving behaviors become more aggressive and unpredictable, interaction uncertainty increases.
In aggressive traffic, although UAMP exhibits a longer travel time, it achieves a substantially lower acceleration cost, resulting in smoother and more stable driving behavior under highly uncertain interactions.

Overall, across all scenarios, UAMP exhibits superior robustness in terms of safety and ride comfort, confirming the effectiveness of incorporating uncertainty in motion planning.

\begin{table}[t]
\centering
\caption{Comparison of performance metrics under varying traffic density levels in the 3-Lane scenario.}
\label{tab:scene_density}
\resizebox{\linewidth}{!}{
\begin{tabular}{ccccccc}
\toprule
\multirow{2}{*}{Density} 
& \multicolumn{3}{c}{$T_{\text{travel}}$}
& \multicolumn{3}{c}{$J_{\text{acc}}$} \\
\cmidrule(lr){2-4} \cmidrule(lr){5-7}
& MVCM & rMAPPO & Ours
& MVCM & rMAPPO & Ours \\
\midrule
Low    & 18.220&15.80 & \textbf{15.500} & 0.668&0.982 & \textbf{0.099} \\
Medium & 14.827&14.70 & \textbf{14.000} & 0.885&0.787 & \textbf{0.454} \\
High   & 22.300&21.88 & \textbf{19.500} & 0.897&0.522 & \textbf{0.399} \\
\bottomrule
\end{tabular}
}
\end{table}

\begin{table}[t]
\centering
\caption{Comparison of performance metrics under varying human driving styles in the 3-Lane scenario.}
\label{tab:hv_style}
\resizebox{\linewidth}{!}{
\begin{tabular}{lcccccc}
\toprule
\multirow{2}{*}{Styles} 
& \multicolumn{3}{c}{$T_{\text{travel}}$}
& \multicolumn{3}{c}{$J_{\text{acc}}$} \\
\cmidrule(lr){2-4} \cmidrule(lr){5-7}
& MVCM & rMAPPO & Ours
& MVCM & rMAPPO & Ours \\
\midrule
Cautious   & 14.611 & 12.90 & \textbf{11.400} & 0.901 & 0.819 & \textbf{0.164} \\
Medium     & 14.827 & 14.70 & \textbf{14.000} & 0.885 & 0.787 & \textbf{0.454} \\
Aggressive & 19.500 & \textbf{18.200} & 20.900 & 1.042 & 0.762 & \textbf{0.174} \\
\bottomrule
\end{tabular}
}
\end{table}

\section{Related Work}
Mixed traffic systems consisting of autonomous vehicles and human-driven vehicles present significant challenges due to the diversity and uncertainty of human driving behavior~\cite{xue2023platoon,chen2021mixed,9246221}. Early research relied on classical parametric traffic and vehicle interaction models to describe vehicle dynamics and behavior~\cite{ghosh2022traffic}. Models such as IDM, OVM, and FVDM characterize traffic evolution using predefined relationships among speed, spacing, and relative motion~\cite{nishinari2014traffic}, offering analytical tractability~\cite{li2023survey}. While these models are computationally efficient and interpretable, they typically assume fixed parameters and deterministic responses, which limits their ability to reflect the variability and uncertainty of human driving behavior in mixed traffic conditions~\cite{guo2020inverse}.

To address these limitations, learning-based methods have been increasingly introduced to capture complex driving patterns. Various data-driven techniques, including neural networks, hidden Markov models, and probabilistic regression approaches, have demonstrated improved accuracy in modeling human behavior compared with traditional parametric models~\cite{di2021survey,morton2016analysis,qu2017modeling,lefevre2014lane}. More recently, reinforcement learning has been applied to enable autonomous vehicles to adapt their decision-making strategies based on observed traffic interactions, without explicitly specifying human driver models~\cite{huang2020learning,chen2023deep,wang2024multiagent,peng2025hierarchical}. These methods improve responsiveness in mixed traffic systems, yet they commonly rely on predictions of human-driven vehicle behavior for downstream decision-making.

Despite their advancements, both traditional and learning-based approaches share a common limitation: predicted behaviors of human-driven vehicles are typically treated as deterministic inputs in planning, while the associated uncertainty is rarely quantified. This omission poses challenges for safety-critical decision-making in complex and dynamic traffic environments.
\section{Conclusion}
Motion planning in mixed-traffic environments relies on accurate predictions of human driving behaviors. However, existing reinforcement learning–based methods often treat predicted intents as deterministic inputs, overlooking the inherent uncertainty in human decision making. To address this issue, we propose Uncertainty-Aware Motion Planning (UAMP), which incorporates proximity-aware intent uncertainty into autonomous decision making. Through Uncertainty-Calibrated Value Learning (UCVL), UAMP enables more cautious and stable control policies. Extensive experiments demonstrate that UAMP improves safety and ride comfort while maintaining competitive traffic efficiency.

\bibliographystyle{named}
\bibliography{ijcai26}

\clearpage
\appendix
\section{Appendix}

\subsection{Different Traffic Scenarios}
\label{app:scenario_settings}

\paragraph{Mixed-Traffic Scenario Design in SUMO}
To evaluate the performance of UAMP under diverse mixed-traffic conditions, we design a set of controlled traffic scenarios in SUMO.
The scenarios are constructed by varying three key factors:
\emph{traffic density}, \emph{autonomous vehicles penetration level}, and \emph{human driving behavior distribution}.

\paragraph{Human Driving Behavior Parameterization}
Human-driven vehicles are modeled using the IDM car-following model in SUMO.
To capture behavioral variability, we consider three representative driving styles, namely aggressive, normal, and cautious, by varying the stochasticity and safety-related parameters of IDM.
Each driving style is characterized by a distinct combination of the imperfection parameter $\sigma$ and the minimum gap $\textit{minGap}$.

\begin{table}[h]
\centering
\caption{Parameterization of human driving behavior styles in SUMO.}
\label{tab:hdv_style_params}
\begin{tabular}{l|cc}
\hline
Driving Style & $\sigma$ & minGap (m) \\
\hline
Aggressive & 0.8  & 1.6 \\
Normal     & 0.5  & 2.0 \\
Cautious   & 0.25 & 2.6 \\
\hline
\end{tabular}
\end{table}

Each human-driven vehicle is assigned a driving style by sampling from a predefined categorical distribution.
By varying the proportions of aggressive, normal, and cautious drivers, we construct traffic flows with different behavior compositions.
Specifically, the balanced setting uses a mixed distribution of 20\% aggressive, 60\% normal, and 20\% cautious drivers.
In the aggressive-dominant scenario, the proportions are set to 60\% aggressive, 30\% normal, and 10\% cautious, while in the cautious-dominant scenario, 10\% of drivers are aggressive, 30\% are normal, and 60\% are cautious.

\paragraph{Traffic Density and autonomous vehicles Penetration Settings}
Traffic density is controlled by the total vehicle generation rate.
We consider three density levels: 800, 1200, and 1600 vehicles per hour, corresponding to low-, medium-, and high-density traffic conditions.
Higher traffic density leads to more frequent interactions among vehicles and increased uncertainty in human driving behaviors.

Instead of specifying autonomous vehicles penetration as a percentage, we fix the absolute number of autonomous vehicles in the traffic flow to better control interaction complexity.
Specifically, we evaluate scenarios with 5, 10, and 20 autonomous vehicles, while the total traffic demand is kept constant.
The remaining vehicles are human-driven vehicles generated according to the specified driving style distributions.

\subsection{Baseline Details}
\label{app:Baseline Details}

To comprehensively evaluate the effectiveness of the proposed UAMP framework, we compare it against a diverse set of baseline methods, including both traditional model-based car-following strategies and modern multi-agent reinforcement learning (MARL) approaches.

\paragraph{Ruled-Based Control Strategy.}

\begin{itemize}[leftmargin=*]
    \item \textbf{MVCM}\cite{wang2021mvcm}: A V2X-based car-following model that incorporates the optimal velocities of multiple preceding vehicles with memory mechanisms. By leveraging information from several leading vehicles, MVCM improves traffic flow stability and enhances disturbance attenuation in mixed traffic environments.
\end{itemize}

\paragraph{Multi-Agent Reinforcement Learning Approaches.}

\begin{itemize}[leftmargin=*]
    \item \textbf{IPPO}\cite{de2020independent}: A fully decentralized MARL method that assigns an independent PPO agent to each autonomous vehicle. Although all agents share the same network architecture, they are trained using only local observations and rewards, without explicit coordination during training.

    \item \textbf{rMAPPO}\cite{yu2022surprising}: A centralized-training decentralized-execution (CTDE) extension of IPPO that incorporates global traffic information during training through a centralized value function. This design enables agents to learn more coordinated behaviors while maintaining decentralized execution.

    \item \textbf{P-DRL}\cite{shi2025predictive}: A hybrid approach that integrates physics-informed behavior prediction models with deep reinforcement learning. By embedding prior knowledge of vehicle dynamics and traffic behaviors into the learning process, P-DRL improves control robustness in mixed-traffic scenarios.

    \item \textbf{MAPPO-PIS}\cite{guo2024mappo}: An intent-sharing MARL framework built upon MAPPO, where agents explicitly exchange high-level driving intentions. This mechanism enhances cooperative decision-making, safety, and overall traffic efficiency among connected and autonomous vehicles.
\end{itemize}

All baseline methods rely on deterministic intent predictions and do not explicitly model uncertainty. To ensure fairness, identical observation and action spaces, as well as consistent training and evaluation settings, are used across all methods.

\subsection{Proof of Value Bias}

\label{app:proof}

To formally justify that directly incorporating uncertain intent predictions into the augmented observation can introduce bias in value estimation, we provide a detailed derivation of the value bias bound.

Consider a surrounding vehicle \( v \) with predicted intent distribution \( p_t^v \) and the corresponding true but unknown intent distribution \( q_t^v \). When the ego agent constructs the augmented observation  
\[
\hat{o}_t^i = (o_t^i, \{p_t^v\}_{v \in \mathcal{N}_i}),
\]
the value function is evaluated based on the predicted distribution rather than the true distribution. This mismatch induces an estimation bias.

The prediction-induced value bias is defined as
\begin{equation}
\Delta V = 
\left|
\mathbb{E}_{p_t^v}[V(\hat{o}_t^i)] -
\mathbb{E}_{q_t^v}[V(\hat{o}_t^i)]
\right|.
\label{eq:bias_def_app}
\end{equation}

Expanding the expectations gives
\begin{equation}
\Delta V
=
\left|
\sum_{a} p_t^v(a) V(\hat{o}_t^i(a))
-
\sum_{a} q_t^v(a) V(\hat{o}_t^i(a))
\right|.
\end{equation}

Rearranging the terms, we obtain
\begin{equation}
\Delta V
=
\left|
\sum_{a} \big(p_t^v(a) - q_t^v(a)\big) V(\hat{o}_t^i(a))
\right|.
\label{eq:bias_expand}
\end{equation}

Applying Hölder inequality yields
\begin{equation}
\Delta V
\leq
\sum_{a}
\left| p_t^v(a) - q_t^v(a) \right|
\cdot
\left| V(\hat{o}_t^i(a)) \right|.
\end{equation}

Since the value function is bounded such that
\begin{equation}
V_{\min} \leq V(\cdot) \leq V_{\max},
\end{equation}
it follows that
\begin{equation}
|V(\hat{o}_t^i(a))| \leq V_{\max} - V_{\min}.
\end{equation}

Therefore, we have
\begin{equation}
\Delta V
\leq
(V_{\max} - V_{\min})
\sum_{a}
\left| p_t^v(a) - q_t^v(a) \right|.
\label{eq:bias_l1}
\end{equation}

The summation term corresponds to the \( \ell_1 \) distance between the predicted and true intent distributions:
\begin{equation}
\|p_t^v - q_t^v\|_1 = \sum_{a} |p_t^v(a) - q_t^v(a)|.
\end{equation}

By the definition of total variation distance,
\begin{equation}
TV(p_t^v, q_t^v) = \frac{1}{2} \|p_t^v - q_t^v\|_1.
\end{equation}

In our method, the uncertainty radius \( \varepsilon_t^i \) is defined such that
\begin{equation}
TV(p_t^v, q_t^v) \leq \varepsilon_t^i.
\end{equation}

Consequently,
\begin{equation}
\|p_t^v - q_t^v\|_1 \leq 2 \varepsilon_t^i.
\end{equation}

Substituting this result into Eq.~\eqref{eq:bias_l1}, we obtain the final bound
\begin{equation}
\Delta V
\leq
2\, \varepsilon_t^i (V_{\max} - V_{\min}).
\label{eq:final_bound}
\end{equation}

This inequality shows that the value estimation bias grows linearly with the uncertainty radius \( \varepsilon_t^i \). Therefore, directly using uncertain predicted intents in the augmented observation can lead to potentially significant value bias, particularly when the prediction uncertainty is large. This result motivates the need for uncertainty-calibrated value propagation in the main method.

\subsection{Model Details}

To summarize the overall learning and decision-making procedure, 
Algorithm~\ref{alg:uamp} presents the complete training pipeline of UAMP,
including proximity-aware uncertainty estimation and uncertainty-calibrated
value learning.

\begin{algorithm}[H]
\caption{UAMP: Uncertainty-Aware Motion Planning}
\label{alg:uamp}
\begin{algorithmic}[1]

\STATE \textbf{Input:} policy parameters $\theta$, value parameters $\phi$;
intent encoder parameters $\psi, \zeta, \eta$;
discount factor $\gamma$, GAE parameter $\lambda$
\STATE \textbf{Output:} trained policy $\pi_\theta$

\FOR{each episode}
    \STATE Initialize rollout buffer $\mathcal{D} \leftarrow \emptyset$

    \FOR{each timestep $t$}
        \STATE Observe physical state $o_t$

        \FOR{each surrounding human-driven vehicle $v$}
            \STATE $\mathbf{h}_t^v \leftarrow \mathcal{F}_\psi(\mathbf{u}_t^v, \mathbf{h}_{t-1}^v)$
        \ENDFOR

        \STATE Construct interaction graph $G_t = (V_t, E_t)$
        \FOR{each $v \in V_t$}
            \STATE $\bar{\mathbf{h}}_t^v \leftarrow 
            \mathcal{G}_\zeta\!\left(
            \mathbf{h}_t^v,
            \{\mathbf{h}_t^u\}_{u \in \mathcal{N}(v)}
            \right)$
            \STATE $p_t^v \leftarrow \pi_\eta(\bar{\mathbf{h}}_t^v)$
            \STATE $H_t^v \leftarrow -\sum_a p_t^v(a)\log p_t^v(a)$
        \ENDFOR

        \STATE Compute ego-centric uncertainty:
        \STATE $\varepsilon_t \leftarrow \sum_{v \in \mathcal{N}_i} w_t(i,v)\, H_t^v$

        \STATE Form augmented observation $\hat{o}_t = (o_t, \{ p_t^v \}_{v \in \mathcal{N}})$

        \STATE Sample action $a_t \sim \pi_\theta(\cdot \mid \hat{o}_t)$
        \STATE Execute $a_t$, observe reward $r_t$ and next observation $\hat{o}_{t+1}$

        \STATE Store $(\hat{o}_t, a_t, r_t, \varepsilon_t)$ in $\mathcal{D}$
    \ENDFOR

    \FOR{each timestep $t$ in $\mathcal{D}$}
        \STATE Compute uncertainty-calibrated TD target:
        \STATE $y_t = r_t + \gamma \min_{\mathbf{Q} \in \mathcal{U}_t} \mathbb{E}_{\mathbf{Q}} \left[ V(\widetilde{o}_{t+1}) \right]$
    \ENDFOR

    \STATE Update $\theta, \phi$ using PPO with targets $\{y_t\}$
\ENDFOR

\end{algorithmic}
\end{algorithm}

\end{document}